\documentclass[10pt,twocolumn,letterpaper]{article}

\usepackage[pagenumbers]{iccv} 
\usepackage{tabularx}
\newcolumntype{U}{>{\arraybackslash}p{1.6cm}}

\definecolor{iccvblue}{rgb}{0.21,0.49,0.74}
\usepackage[pagebackref,breaklinks,colorlinks,allcolors=iccvblue]{hyperref}

\def\paperID{9770} 
\def\confName{ICCV}
\def\confYear{2025}

\title{MeshFleet: Filtered and Annotated 3D Vehicle Dataset for Domain Specific Generative Modeling}

\author{Damian Boborzi\\
University of Augsburg\\
{\tt\small damian.boborzi@uni-a.de}
\and
Phillip Mueller\\
BMW Group\\
University of Augsburg\\
{\tt\small phillip.mueller@bmw.de}
\and
Jonas Emrich \\
TU Darmstadt \\ 
{\tt\small Work done at BMW Group} 
\and
Dominik Schmid \\
University of Augsburg\\
{\tt\small Work done at BMW Group} 
\and
Sebastian Mueller\\
BMW Group
\and
Lars Mikelsons\\
University of Augsburg\\
{\tt\small lars.mikelsons@uni-a.de}
}

\begin{document}
\maketitle
\begin{abstract}
Generative models have recently made remarkable progress in the field of 3D objects. However, their practical application in fields like engineering remains limited since they fail to deliver the accuracy, quality, and controllability needed for domain-specific tasks. Fine-tuning large generative models is a promising perspective for making these models available in these fields. Creating high-quality, domain-specific 3D datasets is crucial for fine-tuning large generative models, yet the data filtering and annotation process remains a significant bottleneck. We present MeshFleet, a filtered and annotated 3D vehicle dataset extracted from Objaverse-XL, the most extensive publicly available collection of 3D objects. Our approach proposes a pipeline for automated data filtering based on a quality classifier. This classifier is trained on a manually labeled subset of Objaverse, incorporating DINOv2 and SigLIP embeddings, refined through caption-based analysis and uncertainty estimation. We demonstrate the efficacy of our filtering method through a comparative analysis against caption and image aesthetic score-based techniques and fine-tuning experiments with SV3D, highlighting the importance of targeted data selection for domain-specific 3D generative modeling. 
\end{abstract}    
\section{Introduction}
\label{sec:intro}
The early-stage engineering design process is a complex and iterative endeavour, characterized by rapid iteration, ideation, feasibility studies, and simulation-driven evaluation. A specific example is automotive design, where engineers and designers rely on digital 3D design representations to explore concepts, evaluate proportions, and ensure compatibility with mechanical and aerodynamic constraints. High-quality 3D models are central to this process and accelerate decision-making, allowing teams to visualize and refine designs before committing to physical prototypes. Generating such structured 3D assets remains a challenge, as manual modeling is both time-intensive and resource-demanding.

Recent advances in 3D generative modeling have shown remarkable potential for creating realistic and diverse 3D content from text or reference images\cite{xiang2024structured, voleti2024sv3d, xu2024instantmesh, liu2023zero1to3, shi2023zero123plus, hunyuan3d22025tencent, yang2024hunyuan3d}. Moving towards such generative models could revolutionize the whole design process in many areas. However, 3D generative models have yet to be widely adopted in industrial design applications as they fall short of generating design representations that exhibit symmetry, geometric consistency, and high levels of detail \cite{alamAutomationAugmentationRedefining2024, 2024SteiningerPotentialsProductDesign, poStateArtDiffusion2023}.

Fine-tuning 3d generative foundation models for engineering design data promises to increase their industrial applicability. Although large-scale 3D datasets like Objaverse-XL \cite{objaverseXL} offer an unprecedented data volume for such endeavors, the availability of datasets in specialized domains is often limited. The performance of these models is highly dependent on the quality and relevance of fine-tuning data \cite{Wallace_2024_CVPR, tan2024vidgen1mlargescaledatasettexttovideo}, and large-scale public datasets contain a significant proportion of noisy or irrelevant samples, necessitating careful filtering and annotation \cite{voleti2024sv3d, xiang2024structured}. 
The general amount of vehicles and vehicle-like objects in Objaverse and Objaverse-XL \cite{objaverse, objaverseXL} is high, with an estimate of more than 20,000 captions from Cap3D \cite{luo2024view, luo2023scalable} and TRELLIS500K\cite{xiang2024structured} depicting cars. However, the actual number of high-quality vehicles which satisfy the requirements for fine-tuning a model for design applications is much lower, the exact number of these objects being unclear. The manual curation of such datasets is time-consuming and expensive. This not only hinders progress in training or fine-tuning 3D generative models for specific domains but also restricts the development of custom methods for conditional control. Here, an analogy to image generation is apparent, where mechanisms such as ControlNet \cite{zhang2023addingconditionalcontroltexttoimage} or Readout Guidance \cite{luo2024readoutguidancelearningcontrol} require training additional adapters on task-specific datasets.

To address these challenges, we propose MeshFleet\footnote{Code is available at: \url{https://github.com/FeMa42/MeshFleet}}, a curated dataset of high-quality 3D vehicle models derived from Objaverse-XL. We define a high-quality vehicle as a single 3D object that is a well-defined car with recognizable make and model, exhibiting detailed contours and representative features. The MeshFleet dataset was constructed in two phases. First, we created a manually labeled subset of Objaverse \cite{objaverse} by automatically identifying potential vehicles through image-based object detection. Each candidate object was then manually annotated with a quality label that reflected its suitability to fine-tune a generative model to produce high-quality vehicle representations.

Second, leveraging this manually labeled data, we trained a quality classifier based on DINOv2 \cite{oquab2024dinov2} features and SigLIP \cite{zhai2023siglip} embeddings. This classifier was designed to automatically identify and filter out low-quality or non-vehicle objects from the larger Objaverse-XL collection. The initial classifier training was followed by an iterative refinement process. This process involved analyzing the description of objects from CAP3D \cite{luo2023scalable} and TRELLIS500K \cite{xiang2024structured} to identify and correct misclassifications, with the corrected samples added to the training data. Furthermore, we incorporated Monte Carlo dropout \cite{gal16_dropout_uncertainty} to estimate the uncertainty of the model, enabling active learning by prioritizing objects with high output entropy for manual review and potential inclusion in the training set \cite{settles2009active}. Our final training data for the classifier consists of 6200 labeled objects from different categories and with varying quality. Only a small subset of these objects belongs to high-quality vehicles. The classifier trained on this data, achieves a $95\%$ agreement with manual labels on a held-out test set.

After finalizing the classifier training, we applied our automated pipeline to process and classify the remaining objects in Objaverse-XL. Specifically, we processed over one million (1,059,727) objects from the Objaverse Alignment \cite{objaverseXL} and TRELLIS500K \cite{xiang2024structured} subsets, both of which contain a diverse collection of high-quality 3D objects. From this processing, we identified and selected 1620 high-quality vehicle models for inclusion in the MeshFleet dataset. A final manual inspection of these filtered vehicles ensured that all included objects met our quality criteria. Beyond the 3D models themselves, MeshFleet includes generated captions and size estimates for each vehicle, providing additional metadata for downstream tasks.

We validate the effectiveness of our labeling and filtering method by fine-tuning SV3D \cite{voleti2024sv3d}, a multiview generative model, on the filtered datasets and comparing the results with those obtained using other filtering strategies. The results demonstrate that finetuning on high-quality objects results in increased quality and multi-view-consistency of the domain-specific generated objects compared to fine-tuning on more, but less relevant data. This underlines that data quality is preferable over data quantity. Finetuning SV3D on significantly fewer, high-quality objects yields better model performance than finetuning on more objects with lower quality. 

Our contributions can be summarized as follows:
\begin{itemize}
    \item A filtered and annotated vehicle dataset extracted from Objaverse-XL, along with embeddings, text descriptions, and vehicle sizes. 
    \item A pipeline for the automated creation of high-quality, domain-specific 3D datasets. Together with a quality-labeled dataset of rendered objects.
    \item A comprehensive evaluation demonstrating the superiority of our approach over existing filtering techniques, highlighting the importance of targeted data selection for 3D generative modeling.
\end{itemize}
\section{Related Work}
\label{sec:related_work}

Existing 3D datasets vary considerably in scale, quality, and the richness of their annotations. ShapeNet \cite{shapenet2015}, while a foundational resource, is limited by its relatively low-resolution models and simplistic textures. More recent datasets, such as ABO \cite{collins2022abo}, GSO \cite{downs2022GSO}, and OmniObject3D \cite{wu2023omniobject3d}, offer improved texture quality but are considerably smaller in scale. Objaverse \cite{objaverse} and Objaverse-XL \cite{objaverseXL} provide unprecedented scale and diversity; however, their heterogeneous quality necessitates effective filtering techniques to extract high-quality subsets \cite{voleti2024sv3d, xiang2024structured}. Manual annotation is inherently expensive and does not scale to the size of these large datasets. Automated annotation methods, such as GeoBiked for images \cite{mueller2024geobikeddatasetgeometricfeatures} and CAP3D for 3D objects \cite{luo2023scalable}, address this scalability challenge by leveraging pre-trained models to generate descriptive text. Other approaches, such as UniG3D \cite{sun2023unig3dunified3dobject} combine different data sets into a uniform multi-modal data representation, including images, text, and mesh information.

Quality assessment of 3D models typically involves evaluating geometric validity, texture fidelity, and semantic consistency. Recent work has explored the use of large pre-trained models for this task, with promising results in the image domain using CLIP embeddings \cite{hessel2021clipscore} and large vision-language models \cite{wu2023qalign}. For 3D quality perception, CLIP embeddings of rendered images have also shown utility. For instance, \citet{xiang2024structured} employed CLIP embeddings of rendered views to estimate aesthetic scores and filter for objects with high aesthetic quality. Combining image-based assessments with 3D representations has been shown to further enhance the performance of quality assessment methods \cite{zhang2022mm}.

Within the domain of 3D vehicles, several specialized datasets have emerged. While both ShapeNet \cite{shapenet2015} and Objaverse \cite{objaverse} include annotated cars, ShapeNet's models often suffer from low resolution and simplistic textures, and the number of high-quality, annotated cars in Objaverse is limited. \citet{chen2024rgmreconstructinghighfidelity3d} introduce a synthetic data generation pipeline for creating 3D car assets under various lighting conditions. DrivAerNet++ \cite{NEURIPS2024_013cf29a} is a multimodal dataset specifically designed for aerodynamic car design, providing 3D meshes, parametric models, aerodynamic coefficients, flow and surface field data, segmented parts, and point clouds for 8,000 car configurations. While DrivAerNet++ offers high-quality data, its primary focus is on engineering and aerodynamic simulation rather than high-fidelity visual representation. In contrast, the 3DRealCar dataset \cite{du20243drealcar} provides scans of 2,500 real-world cars, representing the first large-scale 3D dataset of real car scans with accompanying images and point clouds captured in diverse real-world settings. 3DRealCar offers the advantages and inherent challenges associated with real-world data. It is ideally suited for tasks requiring photorealism or those that must account for the complexities of real-world measurements, including sensor noise, occlusions, and varying visibility conditions. However, for engineering tasks demanding precise shape information, and editable 3D Models, the inherent measurement errors and imprecisions of real-world scanned data may be less desirable than clean, synthetic data.

Our proposed MeshFleet dataset complements these existing resources by providing a collection of curated, high-quality, synthetic vehicle CAD models extracted and filtered from the large-scale Objaverse-XL dataset \cite{objaverseXL}. Unlike 3DRealCar's focus on real-world scans, MeshFleet offers readily editable and customizable 3D models. To the best of our knowledge, MeshFleet is the first dataset to compile such a collection of high-quality, synthetic car models with detailed textures at this scale. We believe that 3DRealCar, DrivAerNet++, and MeshFleet offer complementary resources. 3DRealCar's real-world scans, DrivAerNet++'s parametric car models, and MeshFleet's curated CAD models, each provide unique capabilities for domain-specific research in 3D modeling, generation, and design. The combination of these datasets holds significant potential for enabling experiments in fine-grained 3D generation with domain-specific control and guidance, facilitating research that bridges the gap between synthetic CAD models and real-world car scans.

\section{Data Labeling and Processing}

\begin{figure}
    \centering
    \includegraphics[width=1\linewidth]{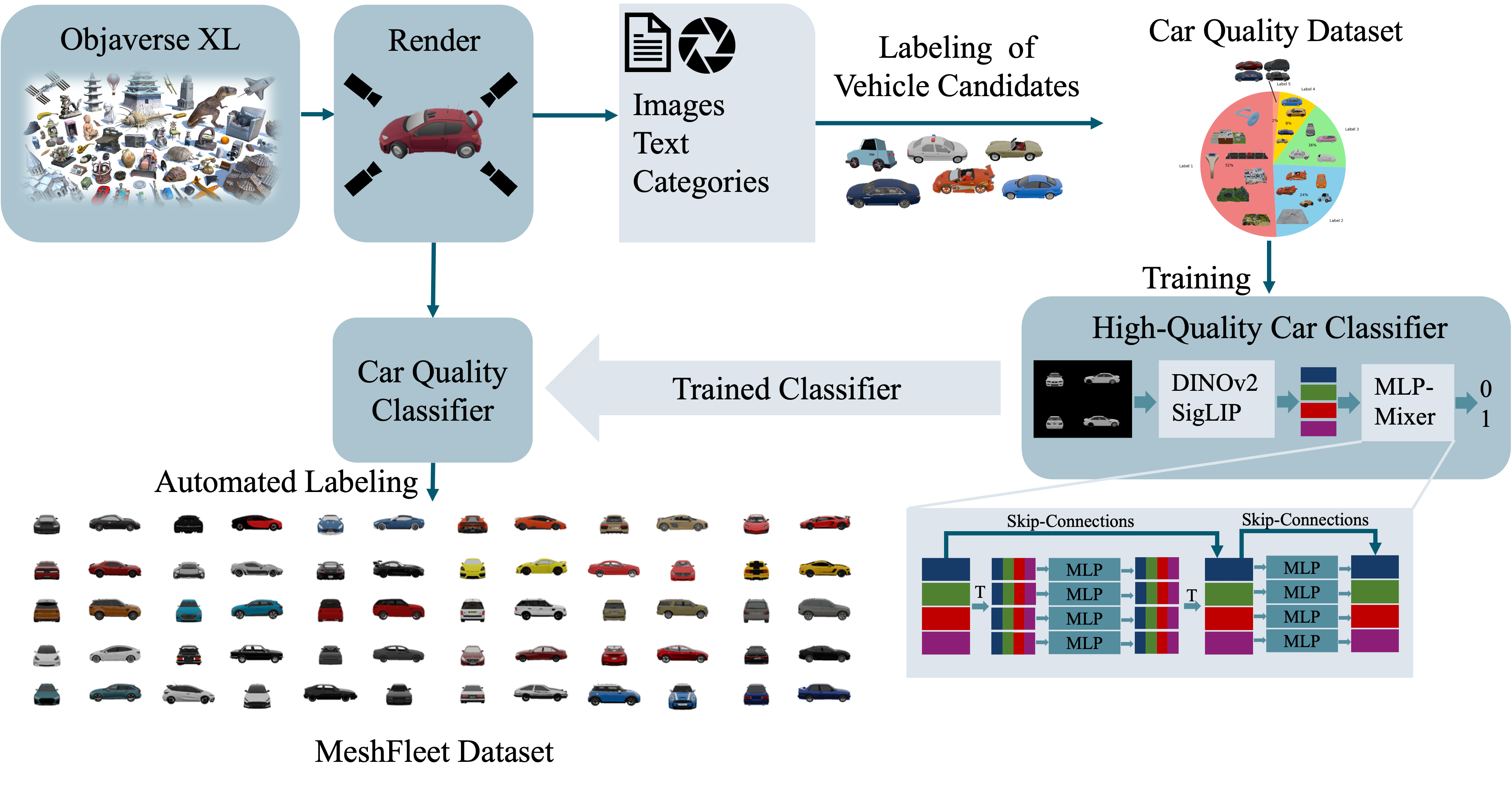}
    \caption{Simplyfied overview of the quality assessment process to generate the MeshFleet Dataset. We render 4 views of each object from high-quality objaverse-XL subsets. We use object detection, clustering and text based filtering to generate a subset of vehicle candidate objects which are subsequently labeled. We then train the High-Quality Car Classifier using the labeled 3D-Car-Quality Dataset. After training we used the trained classifiert to automatically generate the High-Quality Car Dataset which is finally manually reviwed and annotated.}
    \label{fig:pipeline-overview}
\end{figure}

Our proposed pipeline comprises four primary stages: (1) manual quality labeling to create the 3D-Car-Quality Dataset, (2) training of classifier on 3D-Car-Quality Dataset (3) automated quality filtering using the classifier, and (4) iterative refinement of the 3D-Car-Quality Dataset based on textual descriptions and model uncertainty. The automated pipeline operates on a subset of the Objaverse-XL dataset \cite{objaverseXL}, specifically targeting approximately 1,200,000 3D objects from the Alignment Dataset and the remaining objects within TRELLIS500K \cite{xiang2024structured}. For each object, we generate four rendered views using Blender \cite{blender}. These renders are produced at a resolution of 500x500 pixels, capturing the object from equidistant viewpoints along an orbital trajectory at a consistent height and distance after object normalization. From these rendered views, we extract both DINOv2 and SigLIP feature embeddings. We perform dimensionality reduction via Principal Component Analysis (PCA) for the DINOv2 embeddings. These processed embeddings serve as input to the trained quality classifier, which estimates the object's quality label.

\subsection{Manual Quality Labeling}
\label{sec:manual_quality_labeling}

To facilitate the training of a 3D vehicle quality classifier, we created the 3D-Car-Quality Dataset. This dataset was constructed in several stages. Initially, we manually labeled a subset of approximately 4,000 objects from the Objaverse dataset \cite{objaverse}. This initial set comprised objects previously categorized as cars based on LVIS categories, supplemented by additional objects identified as potential vehicles using YOLOv10 \cite{wang2024yolov10}, thereby accelerating the data collection process. We then performed iterative refinement, as detailed in Section \ref{sec:refinement}, incorporating information from image captions and model uncertainty estimates to enhance both the quality and diversity of the labeled data. This iterative process resulted in a final dataset of 6,200 objects from Objaverse-XL \cite{objaverseXL}. Each data point consists of four rendered images of a 3D object, along with a quality label ranging from 1 to 5, reflecting the object's suitability for fine-tuning a 3D generative model specializing in vehicles (see Figure~\ref{fig:proportions-of-each-class} for examples):

\begin{enumerate}
\item \textbf{Unsuitable:} Not a vehicle, extremely low quality, or exhibiting significant rendering artifacts.
\item \textbf{Low-Quality Vehicle:} Recognizable as a car, but with substantial flaws, missing parts, or representing fictional or highly stylized vehicles.
\item \textbf{Average-Quality Vehicle:} A recognizable car, but lacking fine details, overall refinement, and potentially exhibiting minor geometric inaccuracies. May represent specialized vehicle types (e.g., police cars).
\item \textbf{High-Quality Vehicle:} A well-defined car with an identifiable make and model, exhibiting detailed contours and accurately representing characteristic features. 
\item \textbf{Very High-Quality Vehicle:} Exhibiting comprehensive detail, accurate shape representation, high-fidelity texture detail, and complete and accurate feature representation from all viewpoints.
\end{enumerate}

To promote open science and ensure reproducibility, we will publicly release the 3D-Car-Quality Dataset, including both the rendered images and their corresponding quality labels. Providing the rendered images, in addition to the object identifiers and quality scores, is crucial for reproducibility, as the perceived quality can be influenced by the 3D object loading and rendering pipeline.

\subsection{Verification of the Manual Quality Labeling}
\label{sec:verify_manual_labels}

To validate the effectiveness of our manual quality labeling scheme, we conducted a series of fine-tuning experiments using Stable Video 3D (SV3D)~\cite{voleti2024sv3d}, a multi-view diffusion model. SV3D was selected as the foundation model for these experiments due to several key characteristics: its open-source availability, its strong baseline generation capabilities, its moderate computational requirements for fine-tuning, its architecture as an adapted latent diffusion model (which are known to be effective for domain-specific fine-tuning \cite{ruiz2022dreambooth}), and, crucially, its multi-view output representation, which facilitates efficient analysis of object quality and view consistency in image space.

We fine-tuned the UNet component of SV3D using full-parameter optimization, leveraging 21 rendered views per object. Since our focus is vehicle generation, we held out a test set of 12 high-quality car instances for evaluation.  The training datasets were constructed by filtering the manually labeled vehicle data according to the quality label.  Each training subset included all objects with a given quality label and higher. For instance, the "Label 3" subset comprised objects with labels 3, 4, and 5, while the "Label 4" subset contained objects with labels 4 and 5. A hyperparameter search was performed for each label category to identify optimal learning rate schedules (including scheduler-specific parameters) and the number of optimization steps. Table~\ref{table:sv3d_finetune_classes} reports the results for the best-performing run in each category, detailing the training subset size (number of objects), training steps, and the resulting evaluation metrics: CLIP-Score (CLIP-S) \cite{hessel2021clipscore} and mean squared error (MSE) of the generated multi view images using the test set.

\begin{table}[h]
\caption{Fine-tuning subsets with the amount of objects, training steps, epochs, and evaluation metrics. The best scores are bold.}\label{table:sv3d_finetune_classes}
\small
\begin{tabularx}{\columnwidth}{lccccc}
\toprule
 \textbf{Method} & Objects & Steps & Epochs & MSE $\downarrow$ & CLIP-S $\uparrow$ \\
\midrule
\textbf{SV3D}     & 0    & 0     & 0   & 0.0527          & 0.897 \\
\textbf{Label 5}  & 32   & 4000  & 500 & 0.0359          & 0.892  \\
\textbf{Label 4}  & 380  & 8000  & 86  & \textbf{0.0218} & \textbf{0.923}  \\
\textbf{Label 3}  & 1265 & 8000  & 26  & 0.0228          & 0.913 \\
\textbf{Label 2}  & 2474 & 12000 & 20  & 0.0268          & 0.909  \\
\bottomrule
\end{tabularx}
\end{table}

\begin{figure}
    \centering
    \includegraphics[width=1\linewidth]{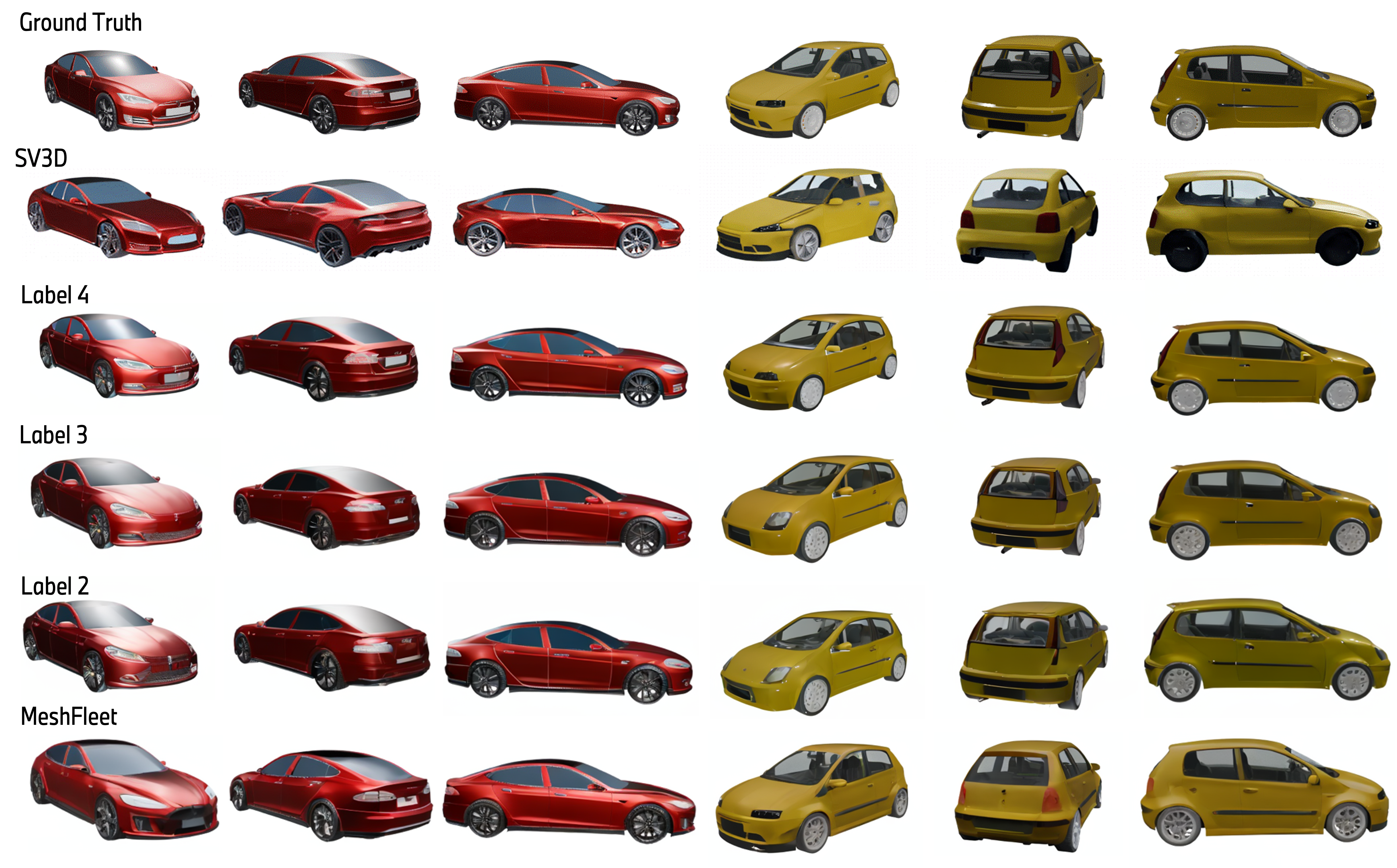}
    \caption{Example views of two vehicles from the Validation set. With the original render (top), SV3D without fine-tuning generated (2nd row), SV3D with \textbf{Label 4} fine-tuning (3rd row), \textbf{Label 3} fine-tuning (4th row), \textbf{Label 2} fine-tuning (5th row), and \textbf{MeshFleet} fine-tuning (6th row).}
    \label{fig:example-generation}
\end{figure}

Although fine-tuning solely on Label 5 data did not improve performance, likely due to the limited number of training instances, the combination of Label 4 and 5 data yielded the best overall fine-tuning results (Table \ref{table:sv3d_finetune_classes}). Surprisingly, incorporating Label 3 objects decreased performance, despite increasing the training set size from 380 (Label 4+5) to 1265 objects (Label 3+4+5) – a more than threefold increase. While most objects with Label 3 are still recognizable as cars (see Figure \ref{fig:proportions-of-each-class}), they exhibit lower fidelity, including reduced geometric detail or representation of specialized vehicle types (e.g., police cars). Further experiments with both extended (1200 steps) and reduced (400 steps) training durations on the Label 3 subset also resulted in lower evaluation metrics. Figure \ref{fig:example-generation} presents example generations from the models fine-tuned on each data subset (more examples are in \ref{supp:addExamples}). Qualitative assessment further underlines the quantitative metrics: the model fine-tuned on Label 4 data produces, in our judgment, the visually superior results. While all fine-tuning runs show improved visual quality compared to the base model, generations from the Label 3 and Label 2 fine-tuned models exhibit less detail and more artifacts compared to those from the Label 4 fine-tuned model. These findings underscore the critical importance of a selective filtering approach that prioritizes data relevance over sheer quantity for optimal fine-tuning. Simply increasing the dataset size with lower-quality or less relevant data proves detrimental. 

\subsection{Comparison of Manual Quality Labels and Aesthetic Scores}
\begin{figure}
    \centering
    \includegraphics[width=1\linewidth]{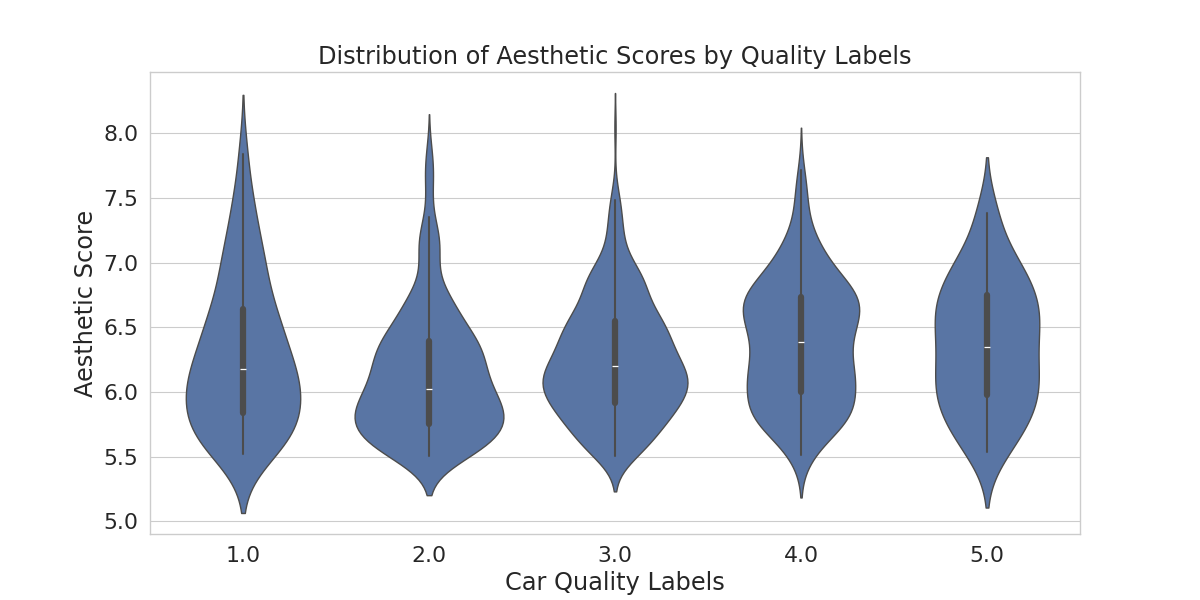}
    \caption{Comparison of the labels from the manual quality labeling (from label 1 to label 5) to the Aesthetic Scores from TRELLIS500K \cite{xiang2024structured}. The plot shows the frequency of aesthetic scores at the different quality labels. We only include data which are described as a car based on the caption from TRELLIS500K.}
    \label{fig:label-vs-aesthetics}
\end{figure}

We further investigate the relationship between our manual quality labels and the aesthetic scores provided in the TRELLIS500K dataset \cite{xiang2024structured}. To isolate car-related objects within TRELLIS500K, we employ a text classification approach using a BART-based large language model \cite{lewis_bart} to identify captions describing cars. This filtering process yields $15,820$ car-related objects from TRELLIS500K (see Section~\ref{supp:trellis500kfiltering} for more details on the text-based filtering), of which $2,833$ overlapped with our manually annotated dataset. Figure \ref{fig:label-vs-aesthetics} presents a comparative analysis of aesthetic score distributions across different quality labels within this overlapping subset. The analysis reveals a weak correlation between aesthetic scores and our assigned quality labels. This discrepancy likely arises because our definition of "high quality" emphasizes domain-specific characteristics such as geometric detail, shape accuracy, and the absence of extraneous objects, criteria not directly captured by a general aesthetic score. Coupled with the fine-tuning results, this observation suggests that filtering solely based on textual descriptions and generic aesthetic scores may be insufficient for specialized fine-tuning tasks requiring high fidelity 3D models.

\subsection{Car Quality Classification}
\label{sec:quality_classifier}

To automate quality assessment for the Objaverse-XL dataset, we developed a binary classification model trained on the 3D-Car-Quality dataset with manually labeled 3D objects. Motivated by the superior fine-tuning performance of SV3D models achieved using high-quality data (Label 4 and 5, as demonstrated in Section~\ref{sec:verify_manual_labels}), our classifier aims to identify objects comparable to these high-quality instances. Therefore, we constructed a binary classification dataset from the manually labeled car subset, assigning a label of $1$ to high-quality cars (original Label 4 and 5) and $0$ to all other instances (original Label 1, 2, and 3).

We represent each 3D object using four rendered views captured from equidistant viewpoints along an orbital trajectory. From these renderings, we extract image features using both SigLIP \cite{zhai2023siglip} and DINOv2 \cite{oquab2024dinov2}. SigLIP provides a sequence of four 768-dimensional feature vectors per object. Recognizing the inherent redundancy in multi-view renderings and the limited size of our manually labeled dataset, we employ Principal Component Analysis (PCA) to reduce the dimensionality of the DINOv2 features. Specifically, we concatenate the DINOv2 features from all four views (initial dimensions: $4 \times 257 \times 768$) into a single vector. We then apply PCA, on the entire labeled dataset, to reduce the dimensionality of the feature vectors to $3072$. This dimensionality was chosen to match the aggregated SigLIP feature dimension, while achieving an explained variance ratio of $0.92$.

We investigated several classifier architectures for our binary quality prediction task. A simple multi-layer perceptron (MLP) operating on concatenated SigLIP features achieved reasonable performance. However, we hypothesized that exploiting the sequential nature of the multi-view features would improve both parameter efficiency and model robustness. Therefore, we compared two sequence processing architectures: Transformer encoders \cite{NIPS2017_transformer} and MLPMixer networks \cite{tolstikhin2021mlpmixer}. For the MLPMixer, we adopted the architecture presented in \cite{tolstikhin2021mlpmixer}, which consists of two types of MLP layers: one applied independently to each feature "patch," and another applied across patches. Instead of image patches, we treated the PCA-reduced DINOv2 features as four patches of 768 dimensions each. These were concatenated with the SigLIP feature sequence ($4 \times 768$) along the patch dimension, resulting in a combined input feature sequence of shape $8 \times 768$. After the MLPMixer layers, the features are average-pooled, and a final MLP layer predicts the object's quality label. We performed hyperparameter optimization for each classifier architecture using an 80/20 train/test split of the manually labeled data. 

We also investigated the influence of the number of rendered views on classifier performance. While using only a single view per object significantly degraded performance, increasing the number of views beyond four provided only marginal gains in accuracy, even with significantly more views (e.g., over 16). Given the increased computational cost associated with rendering and processing a larger number of views, we opted to use four views per object as a compromise between accuracy and computational efficiency. Furthermore, we evaluated the impact of different feature embeddings on the classification accuracy. Individually, both SigLIP and DINOv2 features yielded strong results. However, the combination of both feature sets produced the highest accuracy. Our final binary car quality classifier, utilizing the combined features and the MLP-Mixer architecture, achieved a validation accuracy of $95.0\%$. 

\subsection{Refinement using Captions and Uncertainty Estimation}
\label{sec:refinement}
To improve the accuracy of our initial classification, we incorporated a refinement process utilizing object descriptions from CAP3D and TRELLIS500K. We leveraged text-based classification \cite{lewis_bart} (as in Section~\ref{sec:verify_manual_labels} for the car classification) to identify potential misclassifications by comparing the predicted quality label with the object's described content. For instance, consistent misclassifications of planes and chairs as high-quality cars were identified and rectified by adding these objects to the training set with a label of 0.

\begin{figure}
    \centering
    \includegraphics[width=0.8\linewidth]{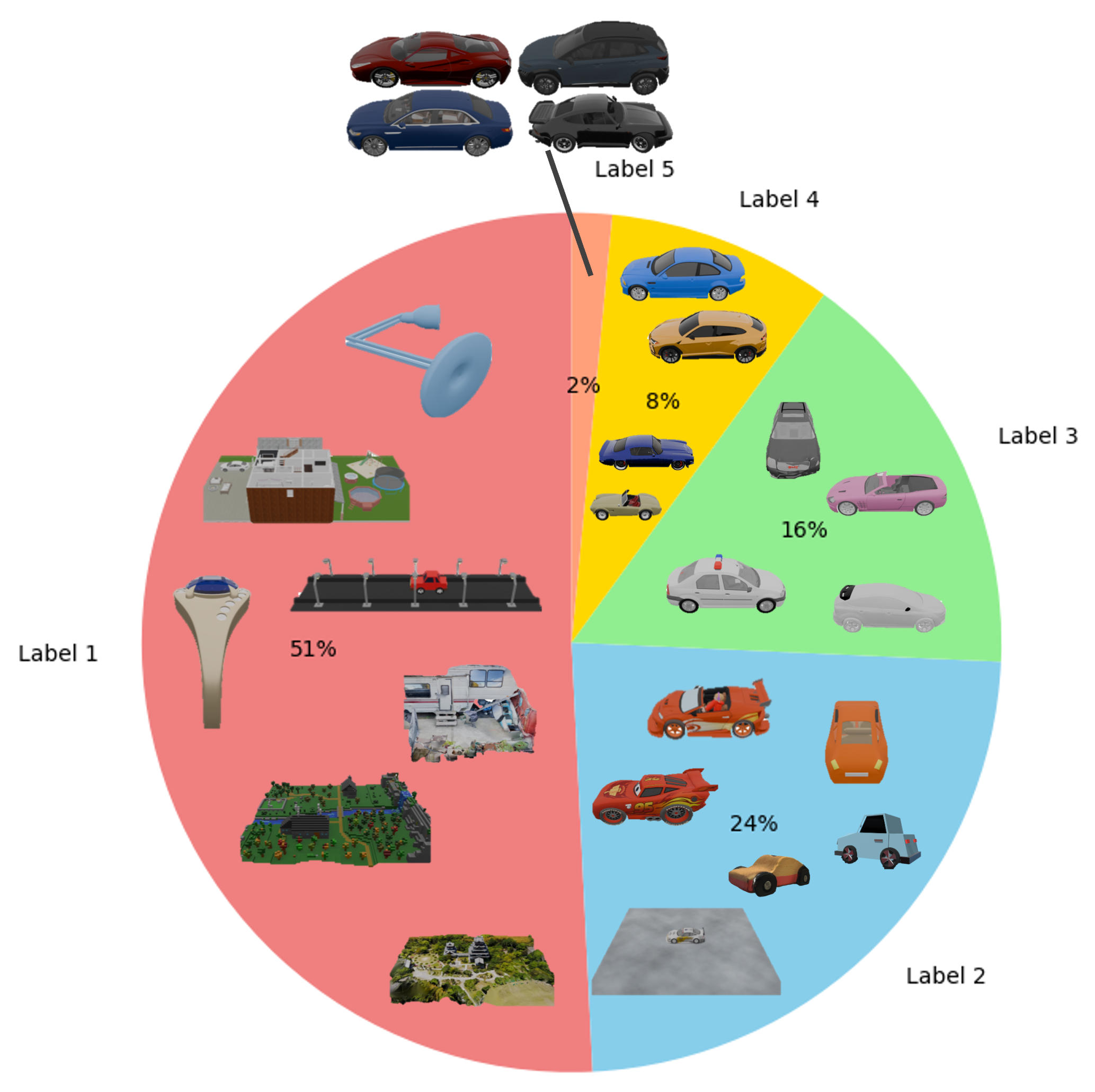}
    \caption{Relative amount of objects in each label categorie for the final dataset we used for training and testing the vehicle classification model. The total amount of objects in the dataset is $6200$. Example objects for each label are shown inside each corresponding section.}
    \label{fig:proportions-of-each-class}
\end{figure}

Furthermore, we employed Monte Carlo dropout \cite{gal16_dropout_uncertainty} to estimate model uncertainty on unseen objects to improve the efficiency of data collection cycles based on active learning techniques\cite{settles2009active}. By activating the dropout layers during inference and performing multiple classifications (we used 500) for each object, we obtained a distribution of predictions. The entropy of this distribution served as a measure of model uncertainty, where high entropy indicates high uncertainty. Objects exhibiting high uncertainty, indicative of ambiguous classifications and the potential to add information to the training set\cite{settles2009active}, were flagged for manual review and, incorporated into the training data. This iterative refinement process, involving four cycles of text-based misclassification detection and uncertainty-guided manual review, resulted in a final labeled dataset of 6200 objects from Objaverse-XL \cite{objaverseXL}. The class distribution of this final dataset is illustrated in Figure \ref{fig:proportions-of-each-class}.

\section{Evaluation and Dataset Validation}


We rendered and processed over one million 3D objects from the Objaverse-XL dataset. Applying the binary classifier described in Section~\ref{sec:quality_classifier}, we initially identified $1,814$ objects as high-quality vehicles. Subsequent manual verification of these classified objects resulted in the selection of $1,620$ objects for inclusion in the final MeshFleet dataset. Representative examples of MeshFleet objects are presented in Appendix ~\ref{supp:examplesmeshfleet}\footnote{Renderings of all MeshFleet objects are available at \url{https://anonymous.4open.science/r/meshfleet-render-6A4A}}.



\subsection{Verification using Fine-tuning Experiments}

The MeshFleet dataset, containing 1620 high-quality vehicle models, is over four times larger than the initial manually labeled subset of Label 4 and 5 objects used for training the classifier. This increase in size is expected to improve the performance of 3D generative models fine-tuned on the domain-specific data. To validate the effectiveness of our automated filtering approach and the quality of the MeshFleet dataset, we conducted a series of fine-tuning experiments using Stable Video 3D (SV3D) \cite{voleti2024sv3d}. These experiments mirror the methodology described in Section \ref{sec:verify_manual_labels}, but utilize the MeshFleet dataset as the training data.

To evaluate the effectiveness of our specialized quality filtering, we conducted a comparative analysis against a filtering approach relying solely on textual descriptions and aesthetic scores. Specifically, we created two subsets from Objaverse-XL, utilizing object descriptions and aesthetic scores from the TRELLIS500K dataset \cite{xiang2024structured}. The first subset, designated CarCaption3K, included all objects filtered based on captions indicative of high-quality, realistic cars (detailed in Section \ref{supp:trellis500kfiltering}). The second subset, CarCaption800, further restricted CarCaption3K by including only objects with an aesthetic score of 6.5 or higher. As with the MeshFleet and manually labeled experiments, we performed a hyperparameter search for each fine-tuning subset. All experiments utilized the same held-out test set of 12 high-quality car instances, ensuring a fair comparison by excluding these instances from all training subsets.

Table \ref{table:sv3d_finetune_meshfleet} presents the results of this comparative analysis. Fine-tuning SV3D with the MeshFleet dataset yielded the highest CLIP-S of 0.925, surpassing even the 0.923 CLIP-S achieved with the Label 4 subset (Table \ref{table:sv3d_finetune_classes}). While the MSE for the MeshFleet fine-tuned model was slightly higher than that of the model trained on the Label 4 subset, the superior CLIP-S indicates improved overall perceptual quality. Fine-tuning with CarCaption3K also improved upon the baseline SV3D performance, but resulted in lower overall metrics compared to fine-tuning with either the manually labeled subsets or MeshFleet. Notably, fine-tuning with the more restrictive CarCaption800 dataset decreased model performance. Manual inspection of objects with high aesthetic scores but low predicted quality revealed a prevalence of toys, fictional vehicles, or partial car models. Although we used more restrictive caption-based filtering, the CarCaption3K dataset still includes these undesirable objects. While we optimized the prompts for the selection of the vehicles in the CarCaption3K dataset, we acknowledge that we did not perform exhaustive prompt optimization or fine-tuning of vision-language models (VLMs) for dirct filtering or generating object descriptions optimized for filtering. Leveraging the labeled data from the 3D-Car-Quality Dataset to perform such optimization represents a promising avenue for future research.

\begin{table}[h]
\caption{Fine-tuning result with the MeshFleet and the caption filtered Dataset. Includes the training steps, epochs, and evaluation metrics. The best scores are bold. Baseline SV3D metrics are in Tabel~\ref{table:sv3d_finetune_classes}.}\label{table:sv3d_finetune_meshfleet}
\footnotesize
\begin{tabularx}{\columnwidth}{Uccccc}
\toprule
 \textbf{Method} & Objects & Steps & Epochs & MSE $\downarrow$ & CLIP-S $\uparrow$ \\
\midrule
\textbf{MeshFleet}     & 1620 & 12000 & 30 & \textbf{0.0229} & \textbf{0.925} \\
\textbf{CarCaption3K}  & 3326 & 20000 & 25 & 0.0318          & 0.902  \\
\textbf{CarCaption800}  & 834 & 8000  & 39 & 0.1390          & 0.862  \\
\bottomrule
\end{tabularx}
\end{table}

\subsection{Dataset Statistics}

The MeshFleet dataset comprises 1620 high-quality 3D vehicle models: 1046 sourced from Sketchfab and 574 from GitHub. The majority of the models (1001) are licensed under a Creative Commons Attribution license. In total, 1112 objects are available under licenses similar to Creative Commons, including variants with non-commercial restrictions, while 503 models have no explicitly specified license.

To enhance the dataset's utility for text-driven tasks, we generated descriptive captions for all vehicles using a specialized prompt with GPT-4o-mini. Although many objects already possess captions from CAP3D or TRELLIS500K, we posit that a consistent set of captions specifically focused on vehicle characteristics (e.g., body style, make, and model) will be more beneficial for downstream applications, such as text-to-3D vehicle generation.

\begin{figure}
    \centering
    \includegraphics[width=1\linewidth]{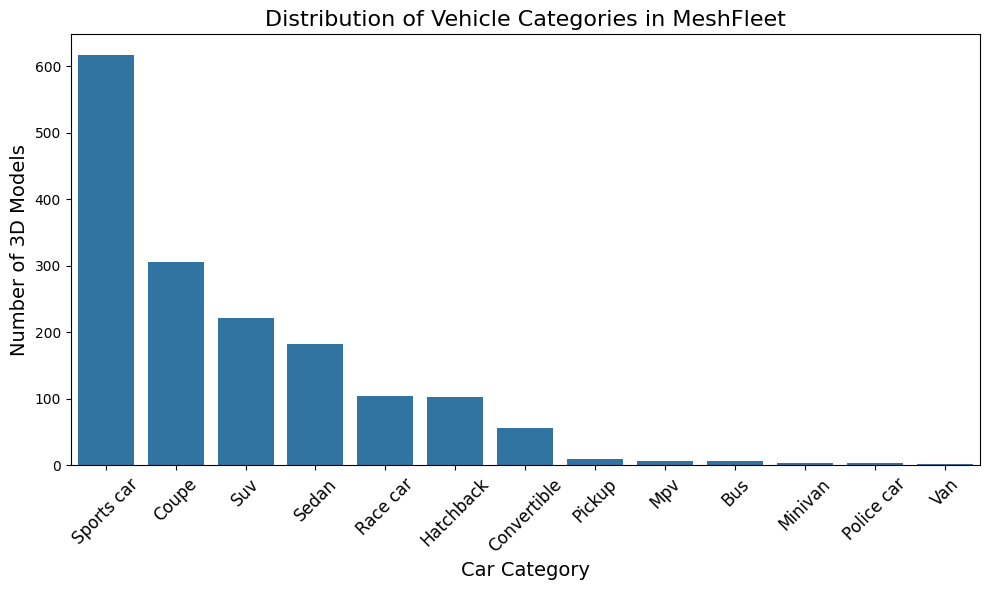}
    \caption{Distribution of vehicle categories within the MeshFleet dataset. The bar chart displays the frequency of each vehicle category (e.g., Sports Car, Coupe, SUV, Sedan) present in the dataset. The x-axis labels indicate the category, and the y-axis represents the number of 3D models belonging to that category.}
    \label{fig:vehicle-types}
\end{figure}

\begin{figure}
    \centering
    \includegraphics[width=1\linewidth]{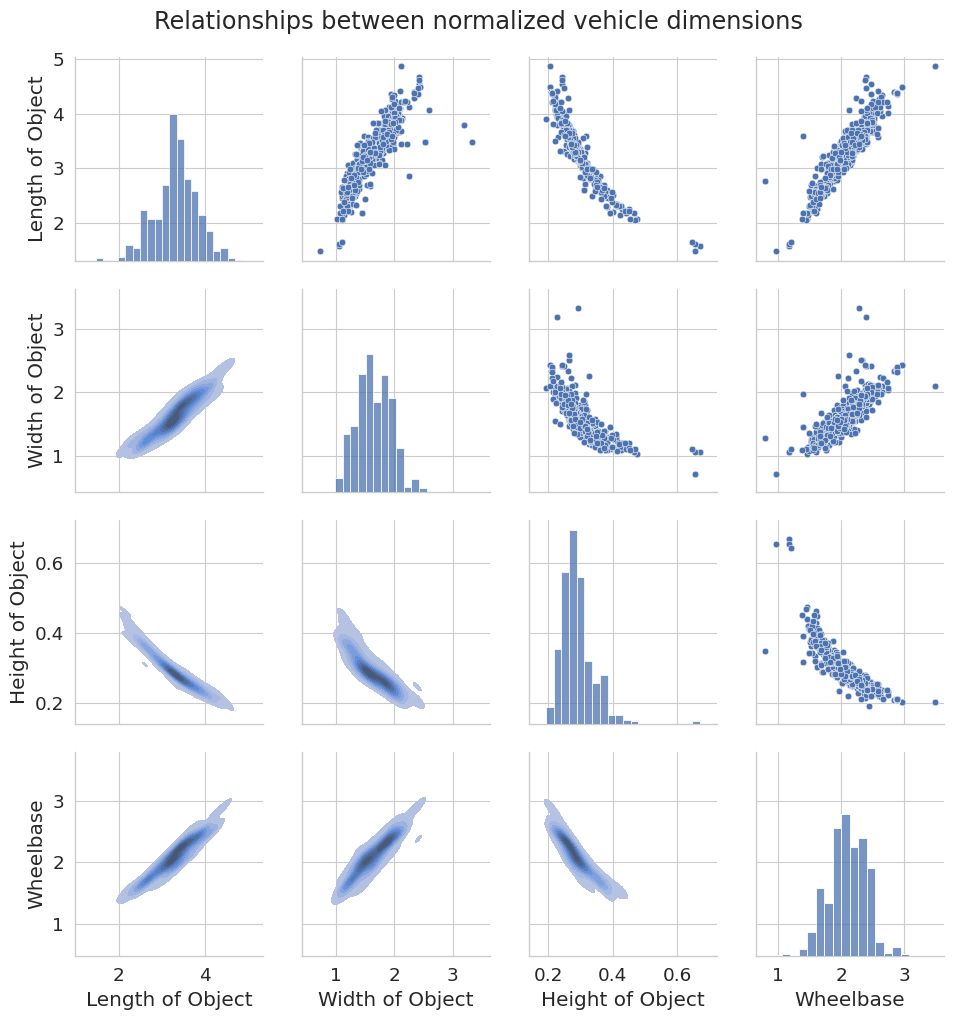} 
    \caption{Pairplot visualizing the relationships between normalized vehicle dimensions in the MeshFleet dataset. The diagonal subplots show the distribution of each dimension (length, width, height, and wheelbase). Off-diagonal subplots show scatter plots (upper triangle) and kernel density estimates (lower triangle) for each pair of dimensions, revealing correlations within the dataset.}
    \label{fig:vehicle-properties}
\end{figure}

In addition to the 3D models and captions, we extracted vehicle-specific metadata for each object, including its category (e.g., sports car, coupe, SUV) and physical dimensions. Figure \ref{fig:vehicle-types} presents the distribution of vehicle categories within the MeshFleet dataset. To facilitate comparisons of vehicle proportions, we derived the length, width, height, and wheelbase of each vehicle from its normalized 3D model. Figure \ref{fig:vehicle-properties} presents a pairplot visualizing the relationships between these normalized dimensions. We anticipate that this detailed metadata will enable further research in 3D generative modeling, facilitating tasks such as shape generation with explicit control over vehicle dimensions (e.g., similar to VehicleSDF~\cite{morita2024vehiclesdf}) and domain-specific conditioning analogous to approaches like ControlNet~\cite{zhang2023addingconditionalcontroltexttoimage} or Readout Guidance~\cite{luo2024readoutguidancelearningcontrol} in the image domain.

\subsection{Additional Data and Preprocessing Details}

To facilitate reproducibility and further research, we will publicly release the code and data used for processing and filtering the Objaverse-XL dataset. While the MeshFleet dataset focuses on high-quality vehicles, reflecting our belief that this focus addresses a critical need in current 3D generative model research, we recognize that the optimal balance between data quality and quantity may vary depending on the specific application.

Therefore, alongside MeshFleet, we will release a comprehensive collection of intermediate data and processing results. This supplementary data includes: rendered images, generated embeddings (SigLIP and PCA-compressed DINOv2), YOLOv10 detections, image captions, classifier outputs (with the uncertainty estimates), and the text classifications for all processed objects from Objaverse-XL. We anticipate that this resource will enable researchers to efficiently filter for domain-specific subsets tailored to their individual needs. For example, by adjusting thresholds on classifier scores and uncertainty estimates, researchers can easily create larger or smaller car-centric datasets with varying quality levels. Many of the processed objects already have associated descriptions from CAP3D \cite{luo2023scalable} or TRELLIS500K \cite{xiang2024structured}. For objects lacking pre-existing captions, we generated descriptive captions using the large version of Florence-2 \cite{xiao2023florence}, utilizing the \texttt{<DETAILED\_CAPTION>} prompt. The provided embeddings, including SigLIP embeddings (useful for rapid clustering and similarity analysis) and the compressed DINOv2 embeddings (used in our classification pipeline), will further expedite downstream tasks.
\section{Discussion}

Our experiments demonstrate the effectiveness of our proposed automated filtering pipeline in creating high-quality, domain-specific 3D datasets. The MeshFleet dataset, generated by this pipeline, significantly outperforms datasets filtered using simpler methods, such as those relying solely on textual descriptions and aesthetic scores, when used for fine-tuning a 3D generative model (SV3D). These results underscore the importance of specialized, task-specific quality assessment for 3D assets, particularly when downstream applications demand high fidelity and geometric accuracy. The strong performance of MeshFleet suggests our approach's suitability for generating domain-specific data tailored to 3D generative fine-tuning.

Despite the promising results, we acknowledge several limitations of our current approach and highlight potential avenues for future research. A primary limitation is the initial dependence on manually labeled data for training the quality classifier. Although general-purpose vision-language models (VLMs) showed promise and substantially reduced our initial manual effort, they did not eliminate the need for further manual refinement. We further mitigated the manual annotation burden through an iterative refinement process incorporating active learning principles (e.g., uncertainty sampling). However, achieving fully automated, high-quality 3D data curation without any manual labeling remains an open challenge.

Another avenue for improvement involves the representation used for quality assessment. Our current pipeline leverages features extracted from 2D renderings (using DINOv2 and SigLIP) from four viewpoints. While this approach represents an effective balance between computational demands and classification accuracy, it may not fully capture all aspects of 3D mesh quality. Incorporating methods that directly analyze the 3D structure, could provide a more comprehensive assessment of geometric and topological quality. Although prior work, such as MM-PCQA \cite{zhang2022mm}, has demonstrated the benefits of combining image features and point cloud data for 3D quality assessment, our preliminary experiments did not reveal a significant improvement in classification accuracy when incorporating point clouds alongside DINOv2 and SigLIP features. Alternatively, 3D-native embedding models, such as the structured latents from TRELLIS \cite{xiang2024structured}, may offer improved performance for 3D similarity and quality analysis due to their direct training on 3D data. However, the computational cost associated with these models must be considered; for instance, the structured latents in TRELLIS are derived from DINOv2 embeddings of 150 renderings, significantly increasing the processing requirements compared to our four-view approach.

\section{Conclusion}

We have presented a novel pipeline for the automated creation of MeshFleet, a high-quality, filtered, and annotated 3D vehicle dataset derived from Objaverse-XL. Our approach combines a quality classifier, trained on a relatively small set of manually labeled data, with iterative refinement leveraging textual descriptions and model uncertainty. Comparative analysis against other existing filtering techniques, and fine-tuning experiments with SV3D, demonstrate the superior performance of our method for creating domain-specific 3D datasets tailored to generative modeling. Notably, models fine-tuned on MeshFleet exhibit improved generation quality within the target domain compared to those trained on datasets filtered solely through automated methods. With the public release of MeshFleet and the 3D-Car-Quality Dataset, we provide a valuable resource to the community and contribute to the advancement of research in 3D generative modeling and large-scale 3D data processing.

{
    \small
    \bibliographystyle{ieeenat_fullname}
    \bibliography{main}
}

\clearpage
\setcounter{page}{1}
\maketitlesupplementary

\section{MeshFleet Further Examples}

\label{supp:examplesmeshfleet}
\begin{figure}[!h]
    \centering
    \includegraphics[width=0.94\linewidth]{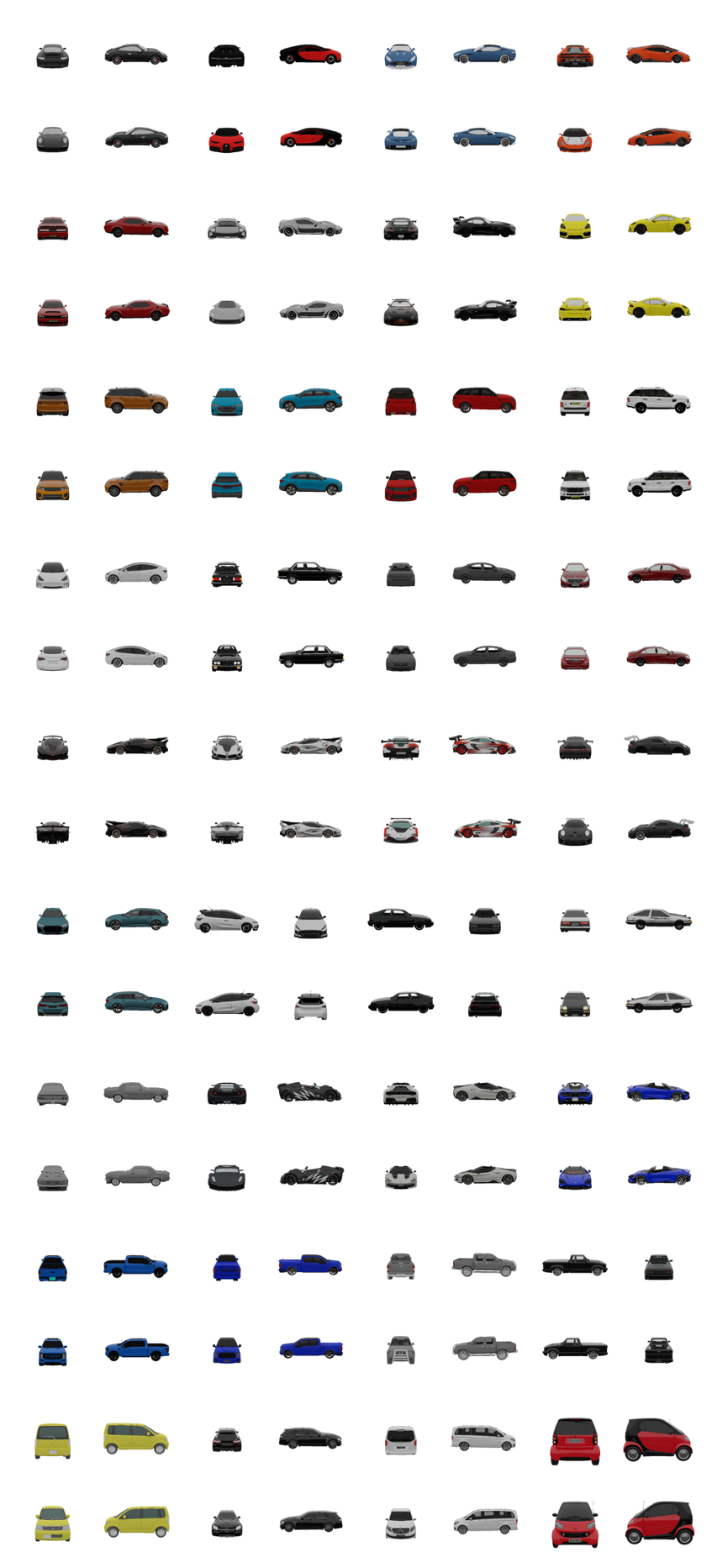}
    \caption{Example renderings from the high-quality vehicle models of the MeshFleet Dataset.}
    \label{fig:examples-meshfleet}
\end{figure}

Figure~\ref{fig:examples-meshfleet} shows several views of example objects from the MeshFleet dataset. We selected a variety of vehicles from different categories to show the diversity of the vehicles. 

\begin{figure}
    \centering
    \includegraphics[width=1\linewidth]{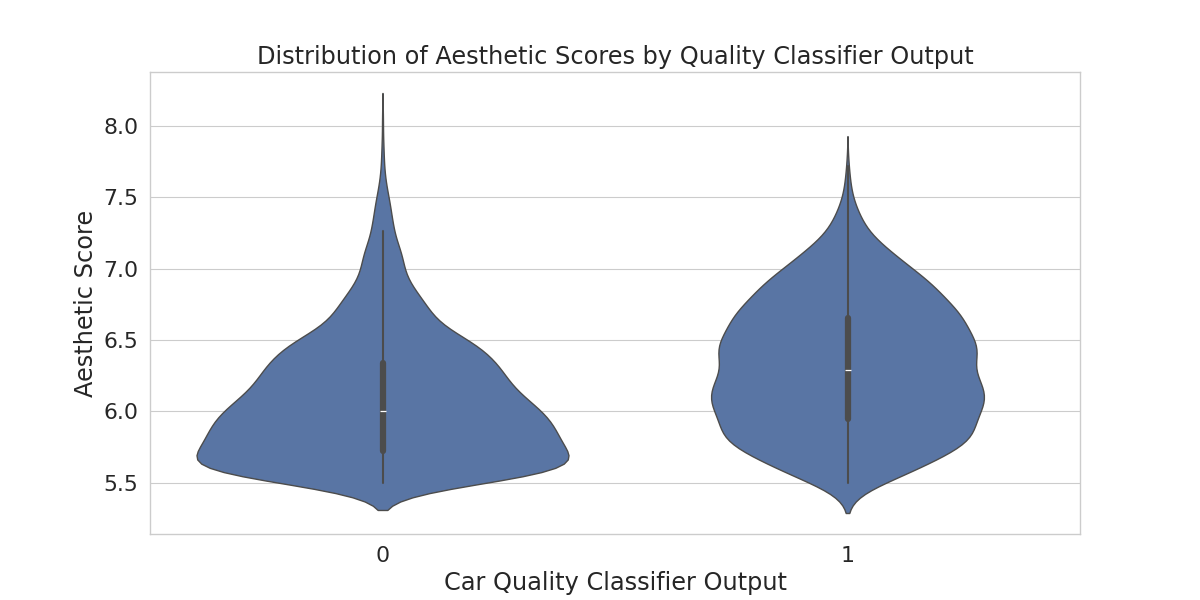}
    \caption{Comparison of predicted quality class labels and aesthetic scores from TRELLIS500K. The plot displays the distribution of aesthetic scores for objects classified as high-quality (suitable for fine-tuning) and low-quality (unsuitable for fine-tuning) by our classifier. Only objects identified as cars based on their TRELLIS500K captions are included.}
    \label{fig:class-vs_aesthetic}
\end{figure}

Figure \ref{fig:class-vs_aesthetic} compares the predicted class labels from our trained quality classifier (high-quality vs. low-quality) with the aesthetic scores from TRELLIS500K \cite{xiang2024structured}, focusing on objects within the CarCaption3K subset. The figure reveals a weak correlation between the predicted quality class and the TRELLIS500K aesthetic scores. Although the mean aesthetic score is slightly higher for objects classified as high-quality, the overall distributions exhibit significant overlap. This indicates that a high aesthetic score in TRELLIS500K does not reliably predict suitability for fine-tuning a generative model on realistic, high-quality car models. 

\section{Vehicle Property Extraction}

For each vehicle in the MeshFleet dataset, we generated descriptive captions using GPT-4o-mini, employing a prompt engineered to generate descriptions detailing the vehicle type, its key characteristics, color, and an assessment of the 3D model's quality.

Beyond the captions, we extracted metadata for each vehicle, including its category and physical dimensions. Vehicle categories were determined by processing the generated captions with a BART-based large language model \cite{lewis_bart}. This model was used to classify captions into the following categories: SUV, Sedan, Hatchback, Pickup, Truck, Minivan, MPV, Coupe, Convertible, sports car, Lorry, race car, police car, and bus.

Physical dimensions (length, width, height, and wheelbase) were derived from the normalized 3D models. Normalization involved scaling each model to fit within a unit cube. Subsequently, 32 views of each normalized vehicle were rendered, and precise 2D bounding boxes were computed for the object in each rendered image, utilizing the transparent background. The side, front, and back views were then identified from these bounding boxes on the basis of their relative dimensions. The wheelbase estimation was performed using Florence-2 \cite{xiao2023florence} with the \texttt{<OD>} prompt to detect tires within the identified side views. All dimensions are reported relative to the normalized vehicle height (a value between 0 and 1), but the original unnormalized height is also provided.

\section{Automated Object Filtering}
\label{supp:trellis500kfiltering}

\begin{figure}[!h]
    \centering
    \includegraphics[width=1\linewidth]{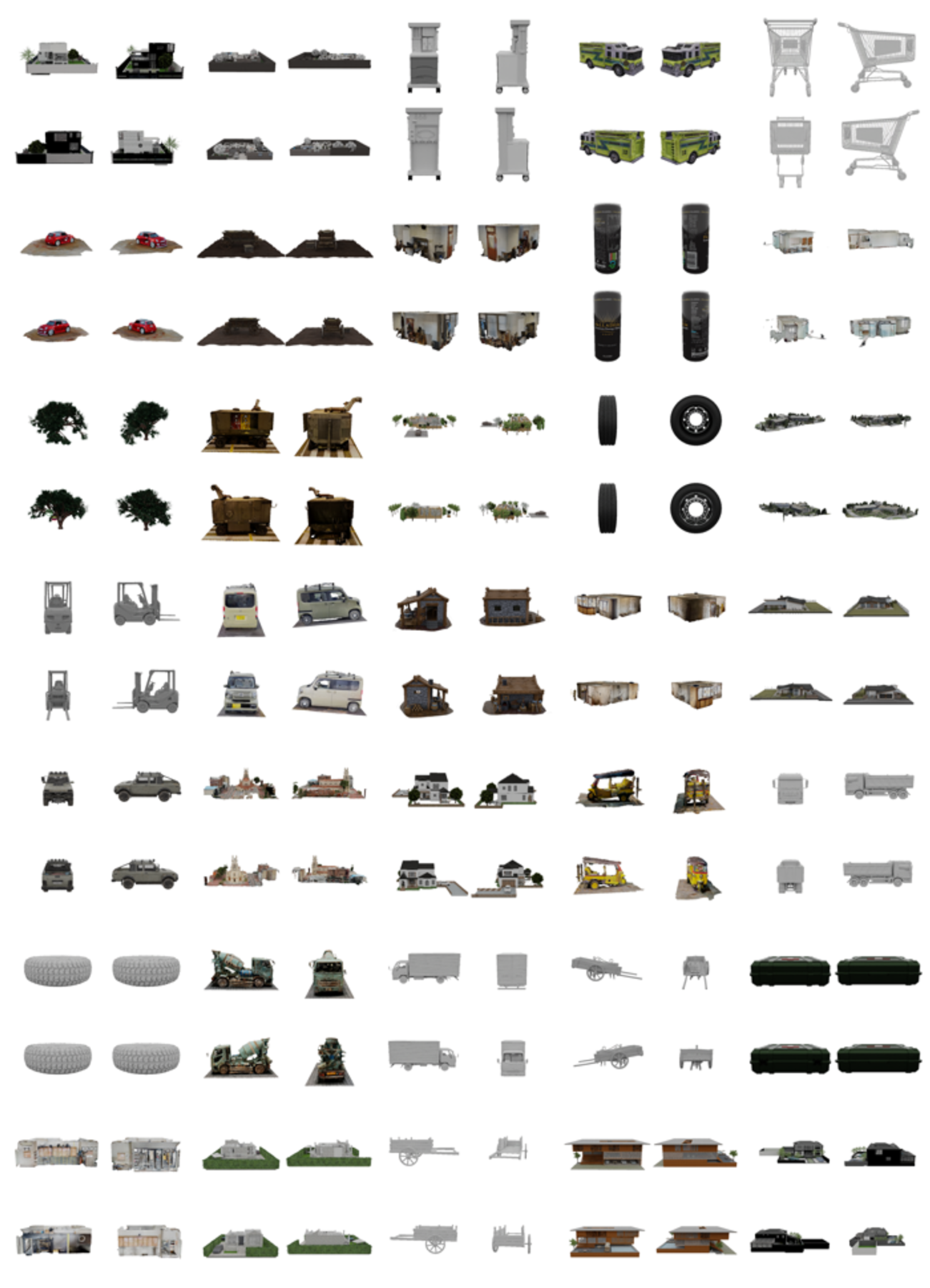}
    \caption{Examples of zero shot image classification of high quality cars using SigLIP.}
    \label{fig:zero-shot-cars}
\end{figure}

To evaluate the MeshFleet dataset, we required a method to automatically identify car-related objects within the larger Objaverse-XL collection. We investigated several approaches, including embedding-based clustering, object detection, and zero-shot image and text classification. Object detection using YOLOv10 \cite{wang2024yolov10} and zero-shot image classification using SigLIP \cite{zhai2023siglip} embeddings produced a significant number of false positives. This is likely due to two factors: (1) many Objaverse objects represent entire scenes containing multiple objects, rather than isolated instances, and (2) both YOLOv10 and SigLIP, like most foundational computer vision models, are primarily trained on photorealistic images, not rendered objects. Figure \ref{fig:zero-shot-cars} presents examples of objects classified as high-quality cars by SigLIP. This initial classification was performed as a two-step process. First, we determined whether the object is a car at all. However, many of these initial detections can also contain simplistic or unrealistic car models. Therefore, in a second step, we attempted to classify the quality of the detected cars, using the following classification categories:
\begin{itemize}
    \item Detailed car model, simplistic car model, partial car model
    \item Realistic car model, toy car, fictional car
    \item High quality car, Low quality car
\end{itemize}

In contrast to image-based methods, text classification based on TRELLIS500K object captions proved to be more promising. We employed a BART-based large language model \cite{lewis_bart} to identify captions describing cars. Filtering based on this criterion yielded 15,820 objects from TRELLIS500K and 20,352 from the combined TRELLIS500K and CAP3D datasets. Given the greater detail in TRELLIS500K captions compared to CAP3D, we focused on the TRELLIS500K captions for subsequent filtering. To further refine this selection and identify realistic and detailed car models, we applied the same categories used in the two-step image classification (detailed above). Filtering for objects labeled as 'Realistic car model', 'Detailed car model', and with a score of at least 0.8 for 'High quality car' resulted in a subset of 3,326 objects. Although manual review of these objects (see Figure \ref{fig:examples-trellis-cars} for examples) confirmed that most correctly depicted cars generally fitting the desired categories, a non-negligible number of partial, simplistic, or non-realistic car models remained.

\begin{figure}[!h]
    \centering
    \includegraphics[width=1\linewidth]{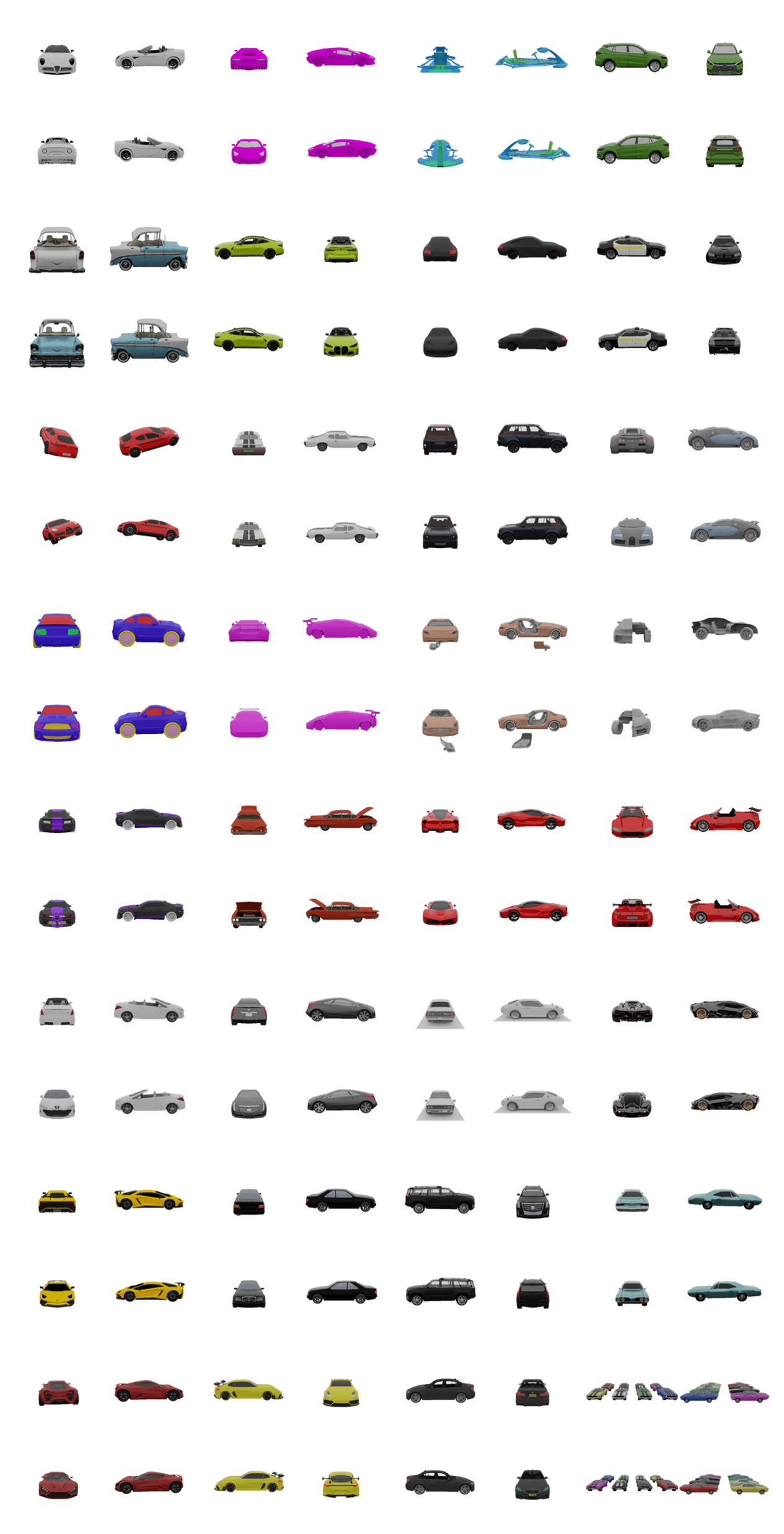}
    \caption{Examples from the dataset generated by filtering Objaverse-XL using object captions from TRELLIS500K.}
    \label{fig:examples-trellis-cars}
\end{figure}

\section{Additional Dataset Assessments}
\label{supp:adddatasets}

\subsection{ShapeNet Car Evaluation}
\label{supp:shapenet_eval}

ShapeNet \cite{shapenet2015} contains a core collection of labeled objects, including a category of approximately 3400 cars. To improve the generalizability of our quality assessment approach, we initially considered incorporating these ShapeNet car models into the 3D-Car-Quality Dataset. We manually labeled a subset of these ShapeNet cars according to the same quality criteria described in Section \ref{sec:manual_quality_labeling}. However, during preliminary experiments with the quality classifier, we observed that including ShapeNet vehicles degraded overall classification performance. We hypothesize that this performance drop is attributable to the relatively homogeneous and simplistic texture representation of ShapeNet models, which typically exhibit a matte appearance.

To mitigate this issue, we experimented with including only the high-quality ShapeNet car models (as determined by our manual labeling) in the 3D-Car-Quality Dataset. This resulted in a subset of 111 high-quality ShapeNet car instances. Incorporating this restricted subset improved the classifier's performance compared to using the full ShapeNet car set.

We further evaluated the trained quality classifier's ability to generalize to unseen data by applying it to the entire labeled ShapeNet car subset. This evaluation can help to assess potential overfitting to the characteristics of the Objaverse-XL objects used in the primary training set. While the classifier had been exposed to the high-quality ShapeNet examples during training, it had not seen the lower-quality ShapeNet instances. Despite this, the classifier achieved a $90\%$ agreement with the manual labels on the complete ShapeNet car subset, using the same configuration as for the Objaverse-XL classification. This result suggests a reasonable degree of generalization beyond the primary training data source, although some domain adaptation may still be beneficial.

\begin{figure}[!h]
    \centering
    \includegraphics[width=1\linewidth]{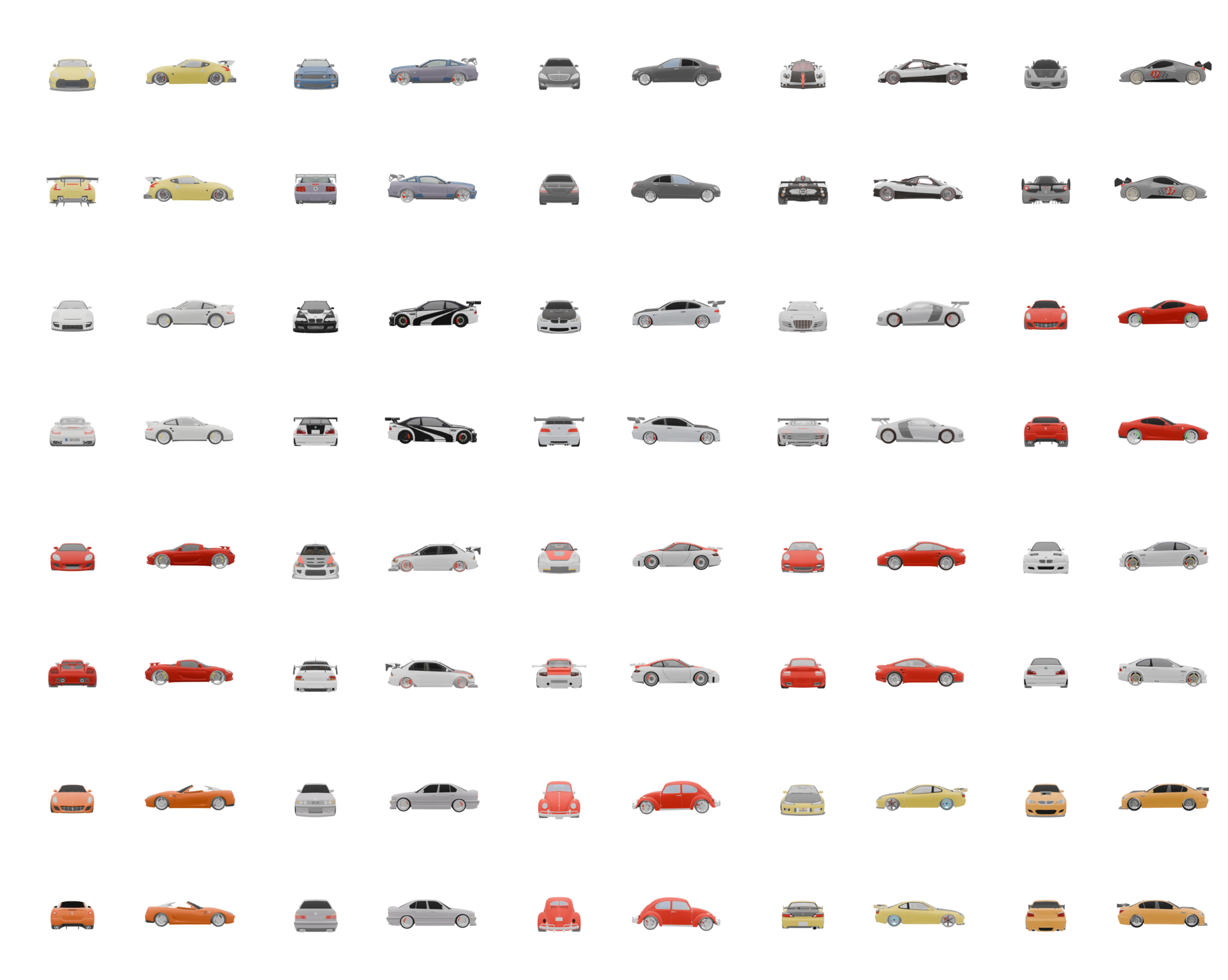}
    \caption{Examples of \textbf{Shapenet} cars classified as \textbf{high quality vehicles}.}
    \label{fig:examples-trellis-cars}
\end{figure}

\begin{figure}[!h]
    \centering
    \includegraphics[width=1\linewidth]{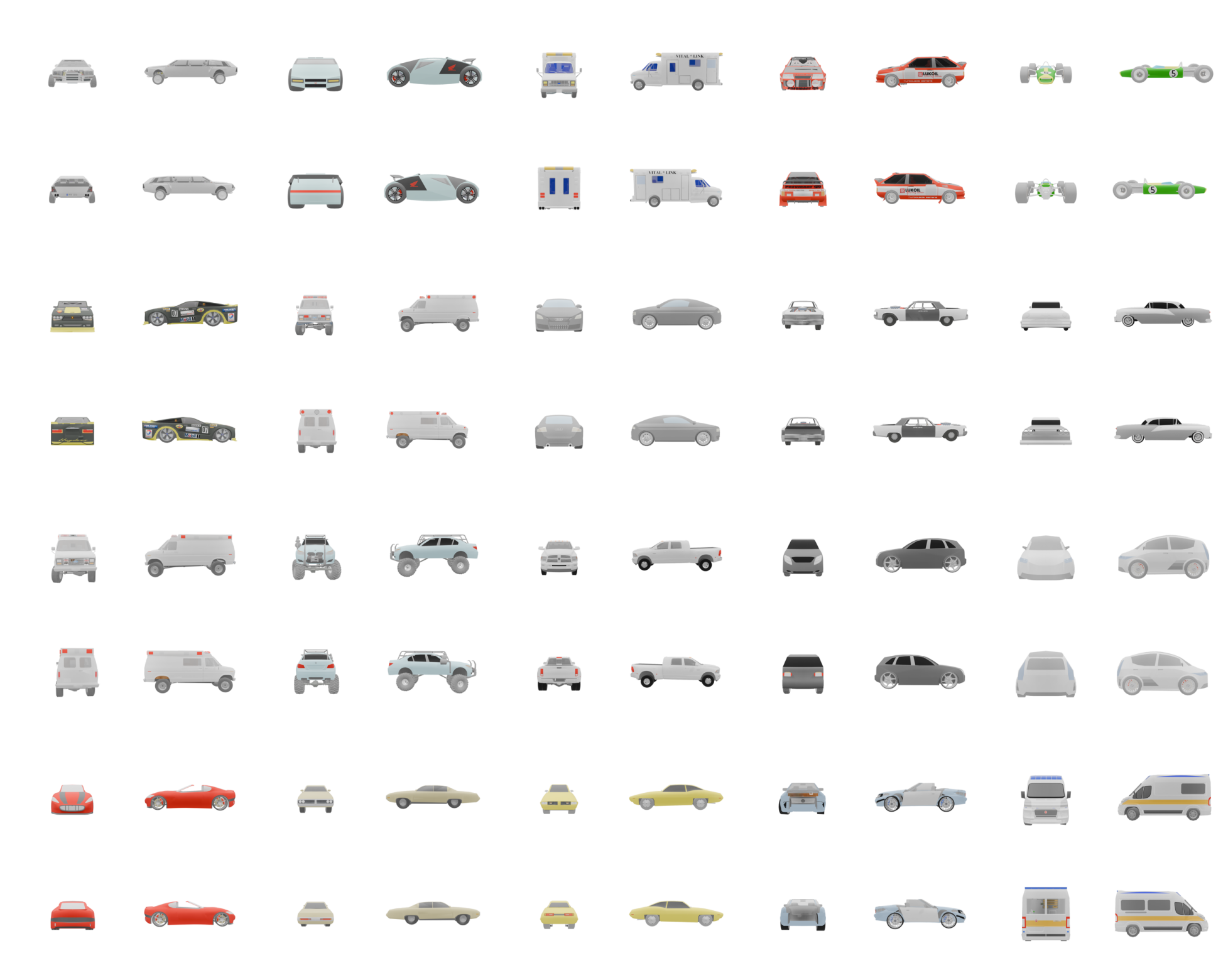}
    \caption{Examples of \textbf{Shapenet} cars classified as \textbf{low quality vehicles}.}
    \label{fig:examples-trellis-cars}
\end{figure}

\subsection{3DRealCar Evaluation}
\label{sec:3drealcar_eval}

To further assess the generalizability of our trained quality classifier and its applicability to real-world car scans, we evaluated it on a subset of the 3DRealCar dataset \cite{du20243drealcar}. We selected 400 vehicles from 3DRealCar and generated rendered images using the same procedure as with Objaverse-XL. As noted previously, Objaverse-XL does contain some real-world scanned car models. However, during the manual labeling process for the 3D-Car-Quality Dataset, we generally classified these scanned models as not meeting the criteria for high-quality vehicles as defined in Section \ref{sec:manual_quality_labeling}. This definition prioritizes detailed, well-defined CAD models, whereas scanned models often exhibit noise, artifacts, or incomplete geometry. Consistent with this prior assessment, the classifier assigned a quality label of 0 (low-quality) to all 400 tested objects from 3DRealCar. This result highlights the classifier's specialization for identifying high-quality synthetic car models and its intended distinction from datasets of real-world car scans.

\section{Additional Example Generations after Fine-tuning}
\label{supp:addExamples}

\begin{figure}
    \centering
    \includegraphics[width=1\linewidth]{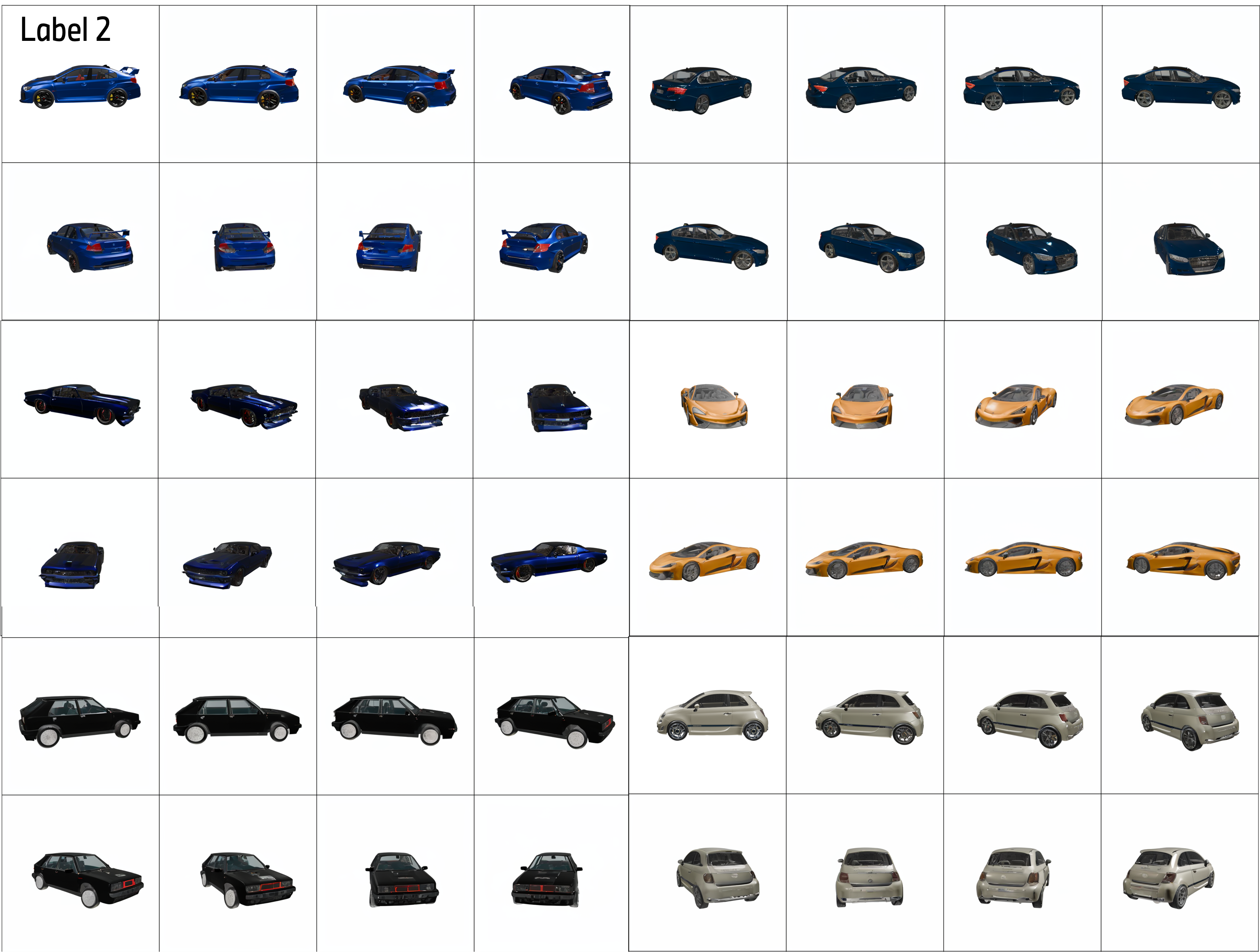}
    \caption{Example view generations of SV3D after finetuning with the \textbf{Label 2} subset. Generated with a single view from the objects of the validation set.}
    \label{fig:example-generations-l2}
\end{figure}

\begin{figure}
    \centering
    \includegraphics[width=1\linewidth]{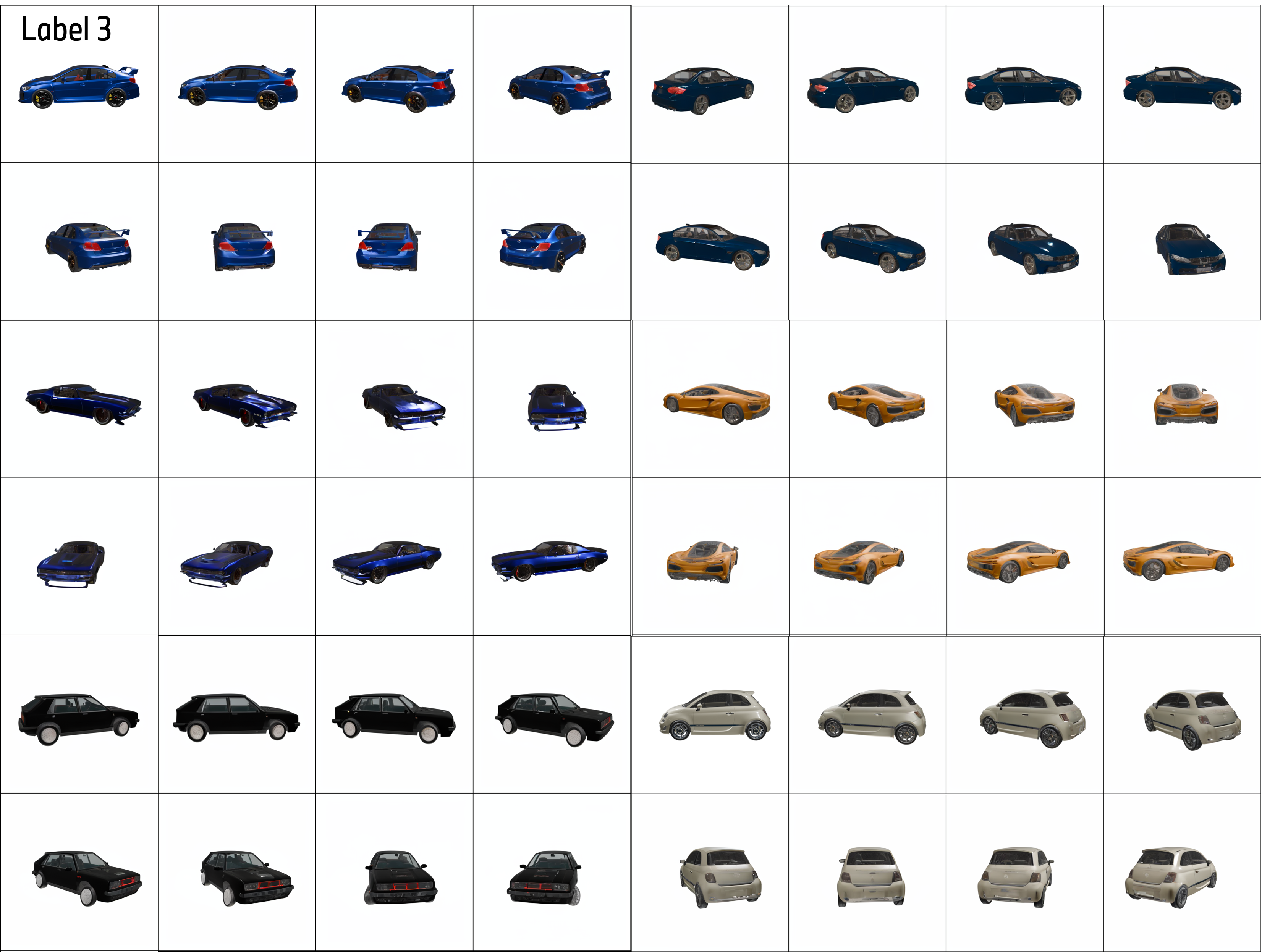}
    \caption{Example view generations of SV3D after finetuning with the \textbf{Label 3} subset. Generated with a single view from the objects of the validation set.}
    \label{fig:example-generations-l3}
\end{figure}

\begin{figure}
    \centering
    \includegraphics[width=1\linewidth]{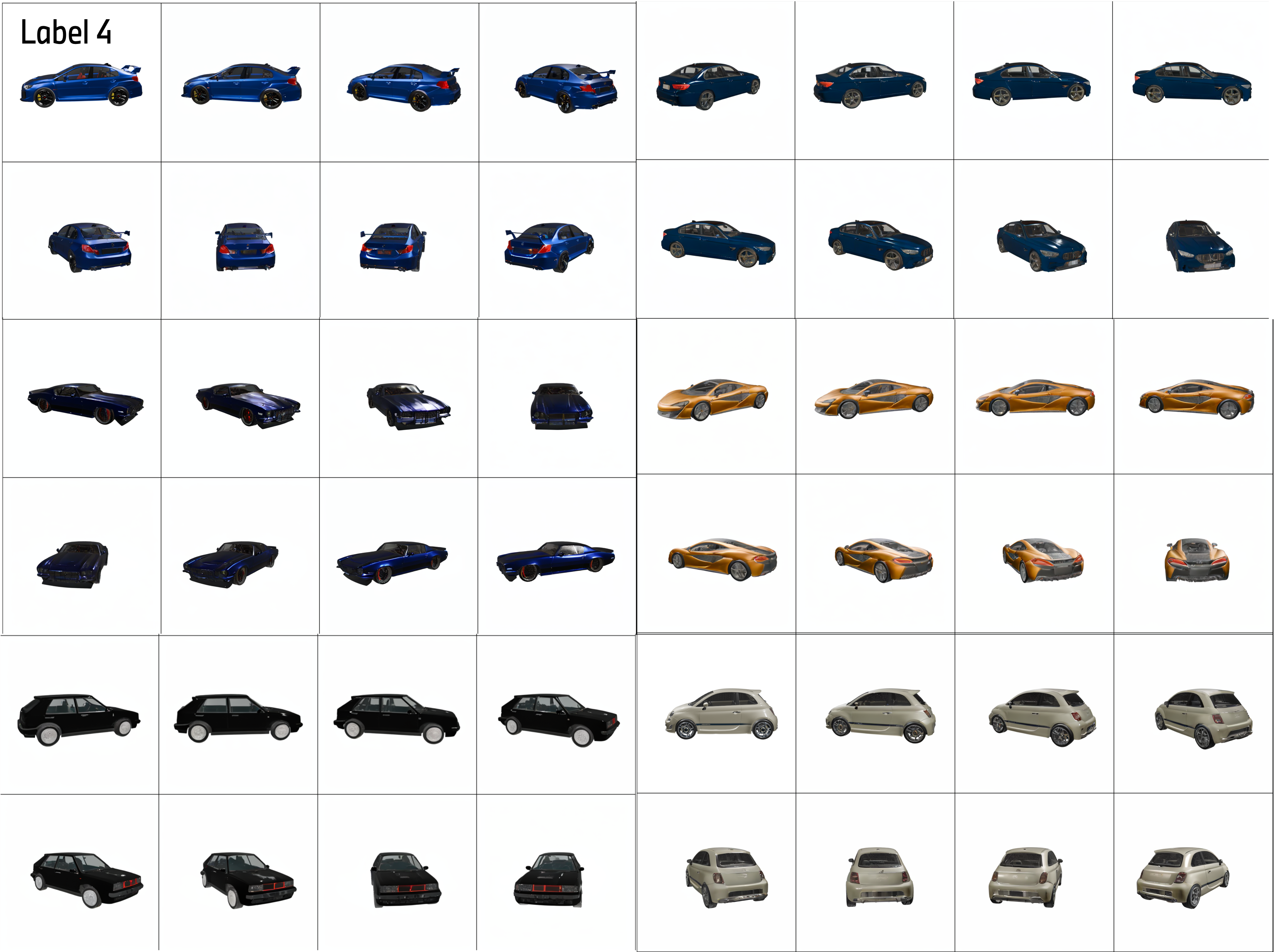}
    \caption{Example view generations of SV3D after finetuning with the \textbf{Label 4} subset. Generated with a single view from the objects of the validation set.}
    \label{fig:example-generations-l4}
\end{figure}

\begin{figure}
    \centering
    \includegraphics[width=1\linewidth]{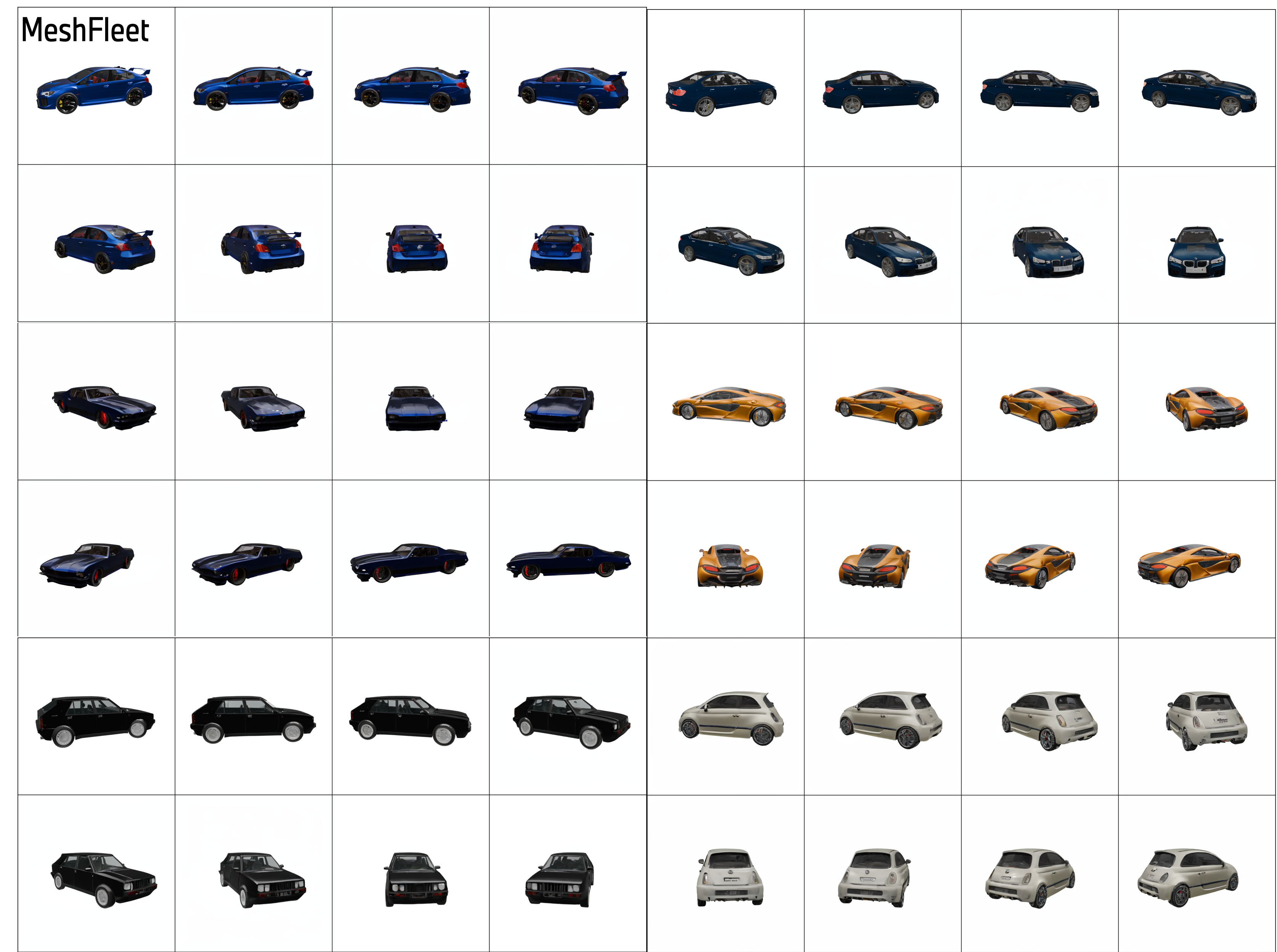}
    \caption{Example view generations of SV3D after finetuning with the \textbf{MeshFleet} dataset. Generated with a single view from the objects of the validation set.}
    \label{fig:example-generations-meshfleet}
\end{figure}

Figures~\ref{fig:example-generations-l2} to \ref{fig:example-generations-meshfleet} show additional examples generated with SV3D after fine-tuning with the respective subsets. All multi-view generations were done with a single view from the objects of the validation set.

\end{document}